\documentclass[letterpaper]{article} 
\usepackage{aaai25}  
\usepackage{times}  
\usepackage{helvet}  
\usepackage{courier}  
\usepackage[hyphens]{url}  
\usepackage{graphicx} 
\usepackage{natbib}  
\usepackage{caption} 
\usepackage{algorithm}
\usepackage{algorithmic}
\usepackage{graphicx}
\usepackage{booktabs}
\usepackage{pifont}
\usepackage{xcolor} 
\usepackage{makecell}
\usepackage{parskip}

\usepackage{multicol}
\usepackage{multirow}
\usepackage{newfloat}
\usepackage{amsmath}
\usepackage{listings}
\DeclareCaptionStyle{ruled}{labelfont=normalfont,labelsep=colon,strut=off} 
\floatstyle{ruled}
\newfloat{listing}{tb}{lst}{}
\floatname{listing}{Listing}
\title{Accelerating Diffusion Models with One-to-Many Knowledge Distillation}
\author{
   Linfeng Zhang$^1$, Kaisheng Ma$^{2*}$
}
\affiliations{
    \textsuperscript{\rm 1}School of Artificial Intelligence, Shanghai Jiao Tong University\\

    \textsuperscript{\rm 2}\footnote{Corresponding author}Institute for Interdisciplinary Information Sciences, Tsinghua University,
}

\usepackage{bibentry}

\begin{document}

\maketitle
\begin{abstract}
  Significant advancements in image generation have been made with diffusion models.
    Nevertheless, when contrasted with previous generative models, diffusion models face substantial computational overhead, leading to failure in real-time generation.
    Recent approaches have aimed to accelerate diffusion models by reducing the number of sampling steps through improved sampling techniques or step distillation. However, the methods to diminish the computational cost for each timestep remain a relatively unexplored area.
    Observing the fact that diffusion models exhibit varying input distributions and feature distributions at different timesteps, we introduce one-to-many knowledge distillation (O2MKD), which distills a single teacher diffusion model into multiple student diffusion models, where each student diffusion model is trained to learn the teacher's knowledge for a subset of continuous timesteps.
    Experiments on CIFAR10, LSUN Church, CelebA-HQ with DDPM and COCO30K with Stable Diffusion show that O2MKD can be applied to previous knowledge distillation and fast sampling methods to achieve significant acceleration.
    Codes will be released in Github.
\end{abstract}
%
\section{Introduction}
Image synthesis is a fundamental challenge in the field of computer vision and representation learning~\cite{pix2pix,cyclegan,gan,vae}. Notably, diffusion models have demonstrated exceptional capabilities in producing high-fidelity and realistic images, surpassing conventional techniques such as generative adversarial models by a substantial margin~\cite{ddpm,iddpm}.

Motivated by their impressive capabilities, diffusion models have been applied in various domains, including text-to-image generation~\cite{latent_diffusion}, image-to-image translation~\cite{diffusion_i2im1,diffusion_distillation_2,diffusion_distillation_4}, and video generation~\cite{diffusion_video_1,diffusion_video_2}, among others.
One significant difference between diffusion models and previous generative models is that diffusion models iteratively perform inference via a denoising network such as UNet or transformer over multiple timesteps. While this property potentially enhances their representational learning capabilities, it also introduces a substantial computational overhead, resulting in increased latency. This challenge has limited the deployment of diffusion models on edge devices and in interactive applications.

The computational costs of diffusion models, denoted as $C_{\text{all}}$, can be roughly approximated as $C_{\text{all}} = T \times C_{\text{single step}}$, where $T$ denotes the number of sampling steps and $C_{\text{single step}}$ signifies the computational cost of inferring the denoising network once. To accelerate diffusion models, recent studies have attempted to decrease the value of $T$ by implementing improved sampling techniques~\cite{ddim} and step distillation methods~\cite{diffusion_distillation_1,diffusion_distillation_2}. Nevertheless, the methods to reduce $C_{\text{single step}}$ have not received extensive attention.

\begin{figure}[t]
    \includegraphics[width=\linewidth]{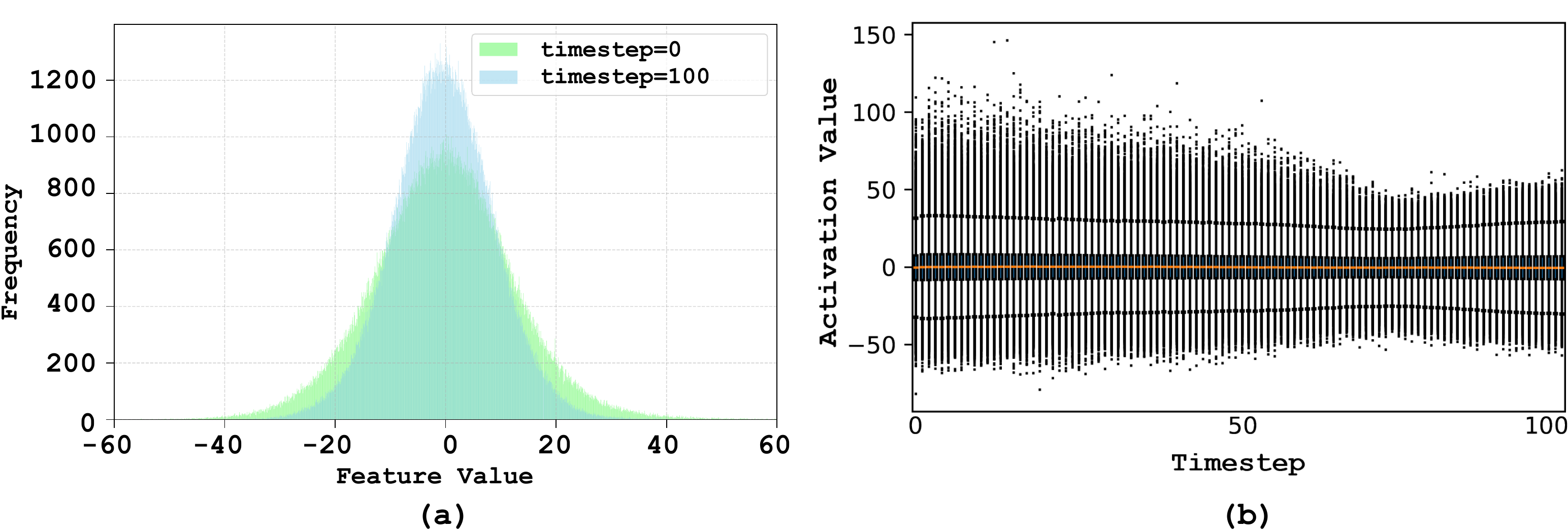}
    \vspace{-0.65cm}
    \caption{Feature visualization of pre-trained diffusion models on CIFAR10  ($T$=100).   (a)~Visualization of feature distribution at timesteps of 0 and 100. (b)~The box plot of feature distribution at all the steps.}
    \vspace{-0.3cm}
    \label{fig:motivation}
\end{figure}


\begin{figure*}[t]
\centering
    \includegraphics[width=\linewidth]{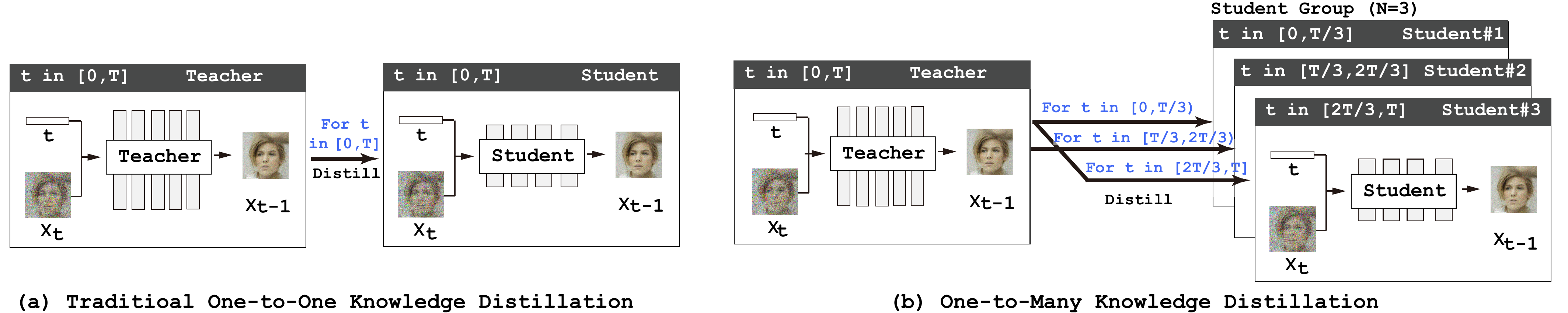}
    \caption{Comparison between traditional one-to-one knowledge distillation and the proposed one-to-many knowledge distillation with three students ($N=3$) in their training period. $T$ indicates the largest timesteps.}
    \label{fig:method}
\end{figure*}
Knowledge distillation (KD), which firstly trains a cumbersome teacher model and then uses the teacher model to train a lightweight student model, has become one of the most effective model compression techniques~\cite{distill_hinton}.
In the original KD, a single student model and a single teacher model are input with the same data, and the student is trained to give a similar output to the teacher model.
By directly applying traditional KD to diffusion models, a single student diffusion model should be trained to mimic a single teacher diffusion model at all timesteps.
For simplicity, we refer to this type of knowledge distillation as ``one-to-one knowledge distillation'' (O2OKD).
Regrettably, our experimental findings reveal that O2OKD 
leads to unsatisfactory performance. To understand the underlying cause, we examine the behavior of diffusion models at different timesteps as follows.

\textbf{\ding{172} The diffusion model exhibits different input distributions across different timesteps.} The 
diffusion model is mathematically formulated as a Gaussian diffusion process which has different distributions in different timesteps~\cite{ddpm}. As the timestep $t$ increases from 0 to its upper limit $T$, the means and variances of the images $x_t$ progressively converge to $\mathbf{0}$ and $\mathbf{I}$, respectively. Notably, since the images at timestep $t$ serve as the input for the diffusion model at timestep $t-1$ during denoising, this indicates variations in the input distribution of the diffusion model.

\textbf{\ding{173} The diffusion model exhibits varying feature distributions at different timesteps.}
As illustrated in Figure~\ref{fig:motivation} and discussed in prior studies~\cite{diffusion_quan1,diffusion_quan2}, diffusion models display different feature distributions at different timesteps. With the progression of timesteps from 0 to its maximum value, the activation range gradually reduces, leading to increased challenges in feature-based knowledge distillation.

\textbf{\ding{174} The diffusion model generates different kinds of information at different timesteps.} As discussed in previous works~\cite{structural_pruning_diffusion},  
diffusion models tend to generate basic contents at larger (noisier) timesteps while generating detailed information at smaller timesteps.

In brief summary, the above observations indicate that an ideal diffusion model should be capable of handling different input distributions, feature distributions, and the generation of different kinds of information.
While this is feasible for the original teacher diffusion model, which contains an ample number of parameters, ensuring sufficient learning capacity, it poses a formidable challenge for the student diffusion model, which has limited parameters. The student model's struggle to match its teacher in handling these complexities ultimately results in the failure of one-to-one knowledge distillation on diffusion models.


Fortunately, we also derive the following observation from Figure~\ref{fig:motivation}(b):
The transition in distributions within diffusion models at different timesteps occurs gradually rather than abruptly. This implies that adjacent timesteps exhibit similar distributions. Hence, it becomes possible to partition the challenge of learning teacher knowledge across all timesteps into multiple sub-tasks, each focusing on learning teacher knowledge within a subset of neighboring timesteps. Given the similarity in distributions among adjacent timesteps, it becomes notably easier for a small student model to address these sub-tasks in comparison to the original task.

\begin{figure}[t]
    \centering
    \includegraphics[width=\linewidth]{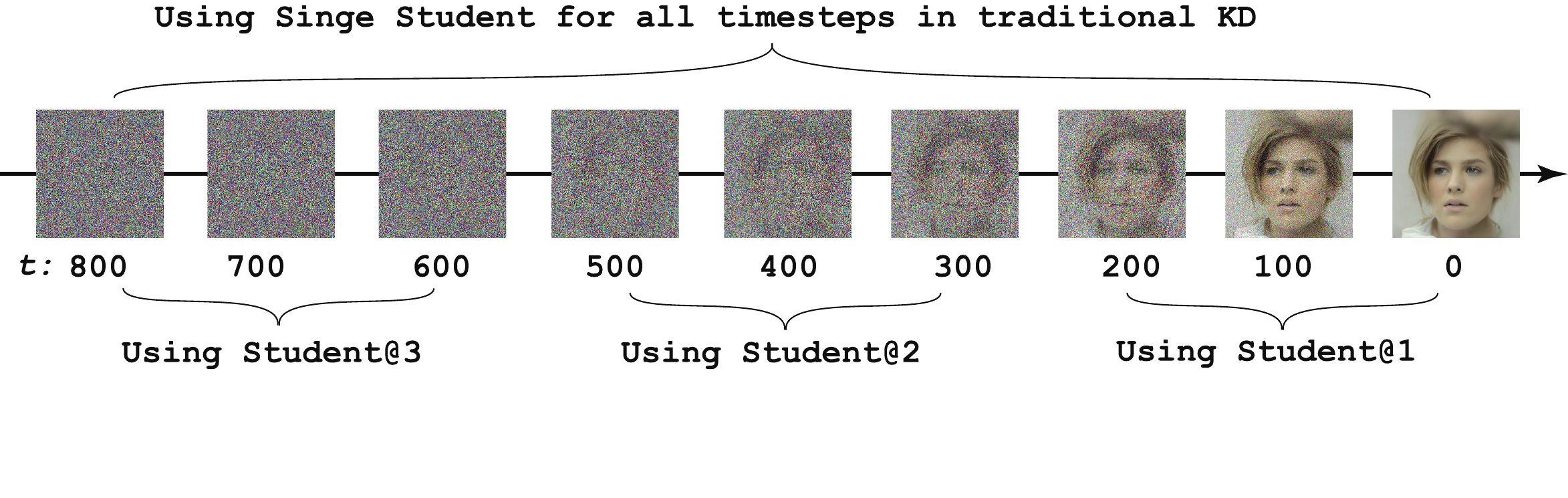}
    \vspace{-1cm}
    \caption{Comparison between traditional one-to-one knowledge distillation and our O2MKD with three students ($N=3$) in the sampling period. $t$ indicates the timestep. }
    \label{fig:sample}
\end{figure}

In light of these insights, we propose ``one-to-many knowledge distillation'' (O2MKD), which aims to distill knowledge from a single teacher model into a group of $N (N>1)$ students. As illustrated in Figure~\ref{fig:method}, during the training phase, each student primarily focuses on learning teacher knowledge within a subset of neighboring timesteps. As shown in Figure~\ref{fig:sample}, in the inference phase, each student is exclusively deployed within its designated timesteps.
For instance, denoting the number of teacher timesteps as $T$, the $i_{th}$ student is exclusively trained to acquire teacher knowledge and utilized for sampling during timesteps $t\in[(i-1)T/N, iT/N]$ in both training and sampling phases, respectively. All students within the group collaborate to generate high-quality images similar to their respective teachers while preserving the same acceleration ratio as one-to-one knowledge distillation.
In O2MKD, since each student is solely responsible for learning the teacher in a subset of timesteps, the learning complexity for each student is significantly reduced. This, in turn, results in superior image fidelity compared to traditional knowledge distillation.



Extensive experimental results demonstrate the effectiveness of O2MKD in various experimental scenarios including CIFAR10, LSUN Church, CelebA-HQ with DDPM, and COCO2013 with Stable Diffusion.  
It's worth noting that O2MKD presents three notable advantages:
\begin{itemize}
    \item \textbf{Compatibility with other acceleration techniques:} O2MKD is designed to alleviate the computational overhead associated with a single step (\emph{i.e.,} the computation of the UNet within the diffusion model) instead of reducing the number of sampling steps. This enables its compatibility with other acceleration techniques such as DDIM, as demonstrated in the discussion section.
    \item \textbf{Compatibility with previous KD:} O2MKD introduces a novel framework involving multiple students and a single teacher, which can be directly applied to most existing knowledge distillation methods.
    \item  \textbf{Beyond Model Compression: } O2MKD can also be utilized in the common diffusion model training to improve the generalization performance, extending its possible utility to the settings beyond model compression.
\end{itemize}
Besides, in the discussion section, we also demonstrate that the additional memory overhead introduced by using multiple students is acceptable, and can be further reduced by using the model merging technique, which merges multiple students into one student after training~\cite{ilharco2022editing}.

This paper brings two insights as the takeaway for readers:
\begin{itemize}
    \item We find that the effectiveness of O2MKD can be explained from the perspective of distillation from a general model (\emph{i.e} the teacher for all the timesteps) into multiple domain experts (\emph{i.e.} multiple students for different timestep ranges), where a hyper-parameter should be utilized to balance the domain knowledge and the general knowledge, as discussed in Figure~\ref{tab:p}.  
    \item As discussed in Figure~\ref{fig:training_loss}, we demonstrate that the supervision from knowledge distillation is significantly more stable than the supervision from traditional training loss of diffusion models. This observation may provide a novel viewpoint for future research in the training methods of diffusion models.
\end{itemize}

\section{Related Work}
\subsection{Diffusion Models}
In the last several years, diffusion models have gained quick
popularity and outperformed other generative models by a clear margin.
Denoising diffusion probabilistic models (DDPM) are firstly proposed to give the basic structure of diffusion models, 
which consists of a forward process for adding noise to images and a backward process for image denoising~\cite{ddpm}.
Then, Nichol~\emph{et al.} improve DDPM by using learnable variances and a cosine noise scheduler~\cite{iddpm}.
After that, latent diffusion models have been proposed to use autoencoder to decrease the latent dimension, which clearly reduces the training and sampling computational overhead~\cite{latent_diffusion}.
Denoising diffusion implicit models (DDIM) are then proposed to formulate diffusion as non-Markovian diffusion processes, which accelerate
the sampling of diffusion models by using fewer sampling steps~\cite{ddim}.
Besides, a few works have tried to accelerate diffusion models by using quantization~\cite{diffusion_quan1,diffusion_quan2},
 pruning~\cite{structural_pruning_diffusion}, and better model scheduling methods~\cite{diffusion_mos_dpm}.
Motivated by the powerful generation ability and the control ability of diffusion models, abundant works have
applied diffusion models to text-to-image generation~\cite{latent_diffusion}, video generation~\cite{diffusion_video_1,diffusion_video_2}, object detection, image super-resolution~\cite{diffusion_sr}, image restoration~\cite{diffusion_restoration}, 3D content generation~\cite{dreamfusion} and so on.

\textbf{Using Different Models in Diffusion Pipelines} Recent works 
have studied the idea of using different models in diffusion pipelines.
EDiff-I is proposed to use experts of denoisers at different timesteps~\cite{balaji2022ediffi}.
Lee~\emph{et al.} propose multi-architecture and multi-expert diffusion models, which use models with different architectures at different timesteps~\cite{Lee2023MultiArchitectureMD}.
Liu~\emph{et al.}  model pipeline, which aims to search a policy to arrange 
multiple pre-trained diffusion models properly~\cite{diffusion_mos_dpm}.


\subsection{Knowledge Distillation}
Knowledge distillation, which aims to transfer knowledge from an over-parameterized teacher model to a lightweight student model,
has become one of the most effective methods in model compression.
Hinton~\emph{et al.} first propose the idea of knowledge distillation which trains the student to mimic the predicted probability distribution of the teachers~\cite{distill_hinton}.
Knowledge distillation has also been utilized for the acceleration of diffusion models.
Progressive distillation has been proposed to align the output of a $T$-step teacher sampler with $T$/2-step student~\cite{diffusion_distillation_2,diffusion_distillation_3}.
Sun~\emph{et al.} further propose to distill the teacher's sharpened feature distribution into the student with a dataset-independent
classifier~\cite{diffusion_distillation_4}.
Moreover, Meng~\emph{et al.} further propose to distill classifier-free guided diffusion models into models that are fast to sample from~\cite{diffusion_distillation_1}.
In conclusion, these methods usually focus  on accelerating diffusion models by reducing the number of sampling steps. How to reduce the computation costs at each timestep, in other words, how to accelerate the UNet in diffusion models, remains a relatively unexplored area.

\section{Methodology}
\subsection{Preliminary}
By denoting the data distribution as $p_{\text{data}}(x)$, a diffusion model $\hat{x}_{\theta}$ with parameters $\theta$
is trained to minimize the weighted mean square error:
\begin{equation}
\mathbf{E}_{t \sim U[0,T], x\sim p_{\text{data}}(x), z_t \sim q(z_t|x)}[\omega(\lambda_t)||\hat{x}_{\theta}(z_t)-x||_2^2],
\end{equation}
where $\lambda_t = \log[\sigma^2_t/\theta_t^2]$ indicates the signal-to-noise ratio
and $\alpha_t$ and $\sigma_t$ indicate the noise scheduling functions.
$q(z_t|x)= \mathcal{N}(z_t;\alpha_tx,\sigma_t^2\mathbf{I})$ and $\omega(\lambda_t)$ is a pre-specified weighting function.

\subsection{Knowledge Distillation}
 \paragraph{Traditional one-to-one knowledge distillation} For simplicity, we denote $f_t=\hat{x}_{\theta_t}$ and $f_s=\hat{x}_{\theta_s}$ as the teacher model and the student model, respectively,
where $\theta_t$ and $\theta_s$ denotes their parameters, respectively.
In naive prediction-based knowledge distillation, a single student model $f_s$ is trained to mimic the generation result of a single teacher model $f_t$, which can be named one-to-one knowledge distillation (O2OKD) and be formulated as 
\begin{equation}
\begin{aligned}
\mathbf{E}_{t \sim U[0,T], x \sim p_{\text{data}}(x), z_t \sim q(z_t|x)} 
& ~\mathcal{L}_{\text{O2OKD}} 
\\= \bigg[ \omega(\lambda_t) ||f_t(z_t) - x||_2^2 
& + \lambda_{\text{kd}} ||f_t(z_t) - f_s(z_t)||_2^2 \bigg]
\end{aligned}
\end{equation}
where $\lambda_{\text{kd}}$ is a hyper-parameter to balance the original training loss (\emph{i.e.}, the first term) and the knowledge distillation loss (\emph{i.e.,} the second term). In this formulation, the student is trained to mimic the predication (generation) result of the teacher, hence it is usually considered as predication-based knowledge distillation. Besides, abundant methods have been introduced to distill the knowledge in teacher features. By denoting the feature encoder of the diffusion model as $\mathcal{E}(\cdot)$, then the one-to-one feature-based knowledge distillation (O2OFKD) can be formulated as 
\begin{equation}
\begin{aligned}
 &\mathbf{E}_{t \sim U[0,T],x \sim p_{\text{data}}(x), z_t \sim q(z_t|x)} 
 ~\mathcal{L}_{\text{O2OFKD}}
\\&= \bigg[ \omega(\lambda_t) ||f_t(z_t) - x||_2^2 
 + \lambda_{\text{kd}} ||g\circ \mathcal{E}_t(z_t) - g\circ \mathcal{E}_s(z_t)||_2^2 \bigg]
\end{aligned}
\end{equation}
where $g$ is a transformation function for the intermediate features such as pooling~\cite{attentiondistillation}, relational extraction~\cite{relational_kd2}, and linear projection~\cite{fitnets}.

\paragraph{One-to-Many Knowledge Distillation (O2MKD)} 
In the aforementioned knowledge distillation methods, the teacher model is distilled into a single student model and thus they can be named as ``one-to-one knowledge distillation'', where the student is expected to handle the same task as the teacher, but with much fewer parameters.
In contrast, our O2MKD, 
we distill the teacher model into a group of $N$ students, which can be denoted as $\mathcal{F}_s = \{f_{s1},f_{s2},\cdots,f_{sN}\}$. 
During training, the $i_{th}$ student is trained to mimic teacher knowledge at timesteps from $(i-1)T/N$ to $iT/N$.
Then, in the sampling period, each image is generated by sequentially inferring $f_{sN},f_{sN-1},\cdots,f_{s1}$ at the corresponding timesteps. 
Intuitively, the training loss of $f_{si}$ can be formulated as 
\begin{equation}
\begin{aligned}
    \label{eq4}
    &\mathbf{E}_{t \sim\textcolor{blue}{U[(i-1)T/N,iT/N]}, x\sim p_{\text{data}}(x), z_t \sim q(z_t|x)} \mathcal{L}_{O2MKD}\\&=[\omega(\lambda_t)||f_t(z_t)-x||_2^2+ \lambda_{\text{kd}}||f_t(z_t) - f_s(z_t)||^2_2],
    \end{aligned}
\end{equation}
where we highlight the difference in the sampling of timesteps. Please note that both prediction-based and feature-based knowledge distillation can be directly utilized in O2MKD by changing the second item. 

\textbf{Trade-off with $\textbf{p}$ in O2MKD}  
In Equation~\ref{eq4}, student $f_{si}$ is exclusively trained to acquire knowledge within its designated timestep period. However, over-specializing the student to a specific range of time steps makes the student not able to benefit from the information at the other timesteps, thereby harming its performance. Our experimental results in the discussion section demonstrate that direct application of Equation~\ref{eq4} during training makes the student totally lose its generation ability in the other timesteps. To achieve a balance between the knowledge of the specific time steps and the global time steps, we introduce the following strategy. For each training iteration, 
the $i_{th}$ student has the possibility of $p$ to be trained with $t\sim U[(i-1)T/N,iT/N]$ with Equation~\ref{eq4}, indicating knowledge distillation for the specific timesteps. Otherwise, it has possibility of $(1-p)$ to be trained with $t\sim U[0,T]$ as the following formulation
\begin{equation}
\begin{aligned}
    \label{eq5}
    &\mathbf{E}_{t \sim U[0, T], x\sim p_{\text{data}}(x), z_t \sim q(z_t|x)} ~~\mathcal{L}\\&=[\omega(\lambda_t)||f_t(z_t)-x||_2^2+ \lambda_{\text{kd}}||f_t(z_t) - f_s(z_t) ||^2_2],
        \end{aligned}
\end{equation}
which indicates knowledge distillation at all the timesteps. 
With a larger $p$, each student is trained to learn more on its designated timesteps.
When $p$ becomes 0, O2MKD degenerates into the common one-to-one knowledge distillation.
Our experimental result shows that a proper $p$ can make students benefit from learning their specific timesteps and achieve good convergence as well.
Another important hyper-parameter in O2MKD is the number $N$ of students in the student group. 
With a larger $N$, each student is trained and sampled for fewer timesteps and a lower FID can be obtained.
However, a larger $N$ also increases the memory usage since more students should be loaded in GPUs. Detailed analysis on $N$, the memory footprint, and the solution to reduce memory footprint is given in the discussion section.

\section{Experiment}

\subsection{Experimental Setting}
\textbf{Models and Datasets}
We primarily evaluate our methods on DDPM using datasets including CIFAR10~\cite{cifar}, CelebA-HQ~\cite{karras2017progressive}, and LSUN Church~\cite{lsun}, and Stable Diffusion with COCO30K~\cite{mscoco}. 
The teacher models employed in our experiments follow their original settings and are available in huggingface. 
We randomly remove some channels from the teacher models and take these models with fewer channels as the students. In other words, the student models have the same architectures and number of layers compared with the teacher but have much fewer channels. The Student-1 in Table~\ref{tab:quan1} and the students in Table~\ref{tab:quan2} have around 30\% channels removed. The Student-2 in Table~\ref{tab:quan1} has around 50\% channels removed. The student for Stable Diffusion is obtained by pruning~\cite{bksdm}. 




\textbf{Implementation}
We follow most default training configurations from the example codes in Diffusers. We set $p=0.5$, $N=4$ or $N=8$, and $\lambda_{\text{kd}}=1.0$ for most experiments. Frechet Inception Distance (FID) is utilized as the metric for quantitative evaluation. A lower FID indicates a higher image fidelity.
We employ DDIM with 100 and 50 steps for sampling on CIFAR10 and other datasets, respectively. 


\setcellgapes[t]{15}
\setcellgapes[b]{15}

\begin{table}[t]
    \centering
      \caption{Experimental results on CIFAR10.}
    \vspace{-0.3cm}
   \renewcommand{\arraystretch}{1.1}
   \resizebox{\linewidth}{!}{\setlength{\tabcolsep}{2mm}{\begin{tabular}{ccclc}
    \Xcline{1-5}{2pt}
    \rule{0pt}{12pt} \multirow{1}{*}{\textbf{Model}}&\multirow{1}{*}{\textbf{Throughput}}& \multirow{1}{*}{\textbf{MACs (G)}}&\multirow{1}{*}{\textbf{KD Method}} & \multirow{1}{*}{\textbf{FID}} \\
    \Xcline{1-5}{2pt}
    \rule{0pt}{12pt}Teacher &10.71 & 6.1 &Training without  KD &4.19 \\ 
    \Xcline{1-5}{0.8pt}
    \rule{0pt}{12pt}\multirow{13}{*}{Student-1}&\multirow{13}{*}{21.56}&\multirow{13}{*}{3.3} &\rule{0pt}{12pt}Training without KD&5.84\\
    \Xcline{4-5}{0.4pt}
    \multirow{14}{*}{{\footnotesize{} }}          &&  & \small~~+ Hinton~\emph{et al.}\rule{0pt}{10pt} KD&5.36\\
                  &&  & \small~~+ Zagoruyko~\emph{et al.} KD&5.51\\
                 &&  & \small~~+ Tian~\emph{et al.} KD&5.44\\
                &&  & \small~~+ Tung~\emph{et al.} KD&5.32\\
       \Xcline{4-5}{0.4pt}
    && &   \small~~+ Ours ($N=4$)\rule{0pt}{10pt} \& Hinton~\emph{et al.} KD&4.73\\
    &&             & \small~~+ Ours ($N=4$) \& Zagoruyko~\emph{et al.} KD&4.81\\
    &&            & \small~~+ Ours ($N=4$) \& Tian~\emph{et al.} KD&4.75\\
    &&            & \small~~+ Ours ($N=4$) \& Tung~\emph{et al.} KD&4.62\\
    \Xcline{4-5}{0.4pt}
    && & \small~~+ Ours ($N=8$) \rule{0pt}{10pt}\& Hinton~\emph{et al.} KD&4.58\\
    &&            & \small~~+ Ours ($N=8$) \& Zagoruyko~\emph{et al.} KD&4.61\\
    &&            & \small~~+ Ours ($N=8$) \& Tian~\emph{et al.} KD&4.58\\
    &&            & \small~~+ Ours ($N=8$) \& Tung~\emph{et al.} KD&4.34\\

    \Xcline{1-5}{0.8pt}
    \multirow{13}{*}{Student-2}&\multirow{13}{*}{39.69}&\multirow{13}{*}{1.4}&\rule{0pt}{12pt}Training without  KD&10.2\\
    \Xcline{4-5}{0.4pt}
    \multirow{14}{*}{{\footnotesize{} }}&&                 & \rule{0pt}{10pt}\small~~+ Hinton~\emph{et al.} KD&8.48\\
    &&             & \small~~+ Zagoruyko~\emph{et al.} KD&8.53\\
    &&             & \small~~+ Tian~\emph{et al.} KD&8.31\\
    &&             & \small~~+ Tung~\emph{et al.} KD&8.35\\
    \Xcline{4-5}{0.4pt}
    &&            & \rule{0pt}{10pt}\small~~+ Ours ($N=4$) \& Hinton~\emph{et al.} KD&  6.42 \\
    &&            & \small~~+ Ours ($N=4$) \& Zagoruyko~\emph{et al.} KD& 6.59 \\
    &&            & \small~~+ Ours ($N=4$) \& Tian~\emph{et al.} KD&6.41 \\
    &&            & \small~~+ Ours ($N=4$) \& Tung~\emph{et al.} KD&6.40\\
    \Xcline{4-5}{0.4pt}
    &          & & \rule{0pt}{10pt}\small~~+ Ours ($N=8$) \& Hinton~\emph{et al.} KD& 6.28\\
    &&            & \small~~+ Ours ($N=8$) \& Zagoruyko~\emph{et al.} KD&6.18 \\
    &&            & \small~~+ Ours ($N=8$) \& Tian~\emph{et al.} KD&6.15\\
    &&            & \small~~+ Ours ($N=8$) \& Tung~\emph{et al.} KD&5.86\\
    \Xcline{1-5}{2pt}
    \end{tabular}}}
    \vspace{-0.4cm}
    \label{tab:quan1}
\end{table}
\begin{figure*}
    \includegraphics[width=\linewidth]{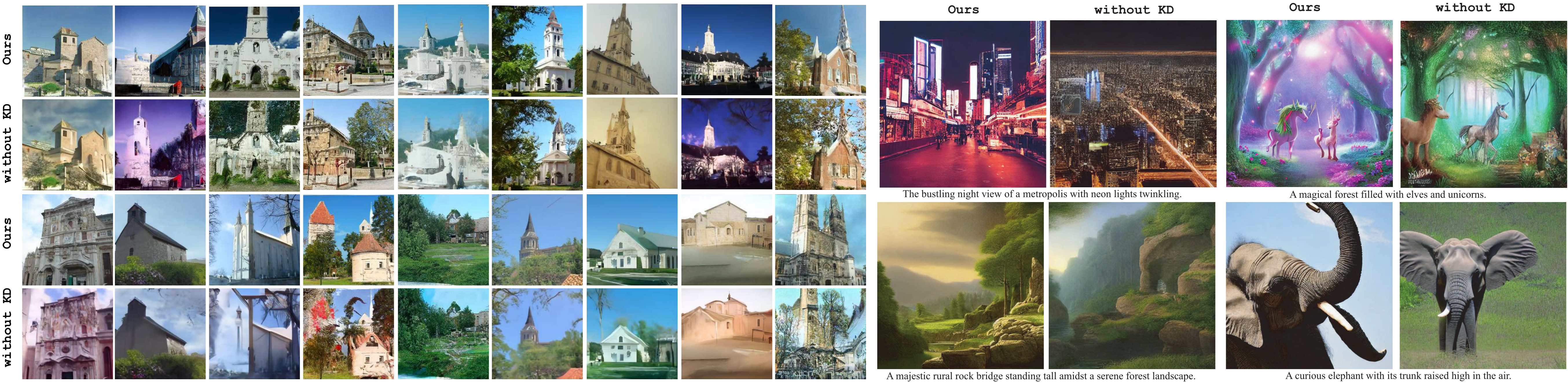}
    \vspace{-0.6cm}
    \caption{Qualitative comparison between the students trained with and without our method. 
    }
    \vspace{-0.5cm}
    \label{fig:lsun_qual}
\end{figure*}
\subsection{Experimental Results}
\paragraph{Quantitative Results}
The quantitative results of our O2MKD are presented in Table~\ref{tab:quan1}, Table~\ref{tab:quan2} and Table~\ref{tab:quan3}, respectively.
Our observations are as follows:
    (1) On CIFAR10, O2MKD results in a 1.8$\times$ acceleration with only a 0.18 FID improvement. On average, the students trained with O2MKD exhibit a 2.57 lower FID compared to students trained without knowledge distillation, demonstrating its effectiveness.
    (2) Applying O2MKD in conjunction with the four knowledge distillation methods yields a 1.46 FID reduction compared to applying these methods in the traditional one-to-one knowledge distillation framework, indicating the generalization ability of our approach across various knowledge distillation methods.
    (3) It is observed that by using more students (\emph{i.e.,} a larger $N$), the FID can be further reduced. For instance, on CIFAR10, using eight students leads to greater FID reduction compared to using four students on both the two students, indicating that the effectiveness of our method can be further enhanced by using more students.
    (4) O2MKD consistently reduces FID on more challenging datasets and tasks. On average, O2MKD leads to a 24.12 and 14.82 FID reduction when compared to students trained without knowledge distillation on LSUN Church and CelebA-HQ, respectively. On COCO30K, compared with the student trained by traditional KD, our method leads to 2.91 decrements in FID, and 4.23 and 0.0218 improvements on IS and CLIP Score, respectively, indicating a significantly better generalization quality. 

\begin{table}[t]
    \centering
      \caption{Experimental results on LSUN Church. }
   \vspace{-0.3cm}

   \renewcommand{\arraystretch}{1.1}
   \resizebox{\linewidth}{!}{\setlength{\tabcolsep}{1mm}{\begin{tabular}{cccllccccll}
    \Xcline{1-5}{2pt}
    \rule{0pt}{12pt} \multirow{1}{*}{\textbf{Model}}&\multirow{1}{*}{\textbf{Throughput}}& \multirow{1}{*}{\textbf{MACs (G)}} &\multirow{1}{*}{\textbf{KD Method}} & \multirow{1}{*}{\textbf{FID}} \\
    \Xcline{1-5}{2pt}
    \rule{0pt}{12pt}Teacher&1.47&248.68&Training without  KD &10.60\\ 
    \Xcline{1-5}{0.8pt}
    \multirow{10}{*}{Student}&\multirow{10}{*}{3.21}&\multirow{10}{*}{119.20} &\rule{0pt}{12pt}Training without KD&35.86\\
    \Xcline{4-5}{0.4pt}
                   &&  & \small~~+ Hinton~\emph{et al.}\rule{0pt}{10pt} KD&27.32\\
                  &&  & \small~~+ Zagoruyko~\emph{et al.} KD&28.22\\
                 &&  & \small~~+ Tian~\emph{et al.} KD&26.80\\
                &&  & \small~~+ Tung~\emph{et al.} KD&24.09\\
       \Xcline{4-5}{0.4pt}
    &&&   \small~~+ Ours ($N=4$)\rule{0pt}{10pt} \& Hinton~\emph{et al.} KD&14.72\\
    &&             & \small~~+ Ours ($N=4$) \& Zagoruyko~\emph{et al.} KD&15.40\\
    &&            & \small~~+ Ours ($N=4$) \& Tian~\emph{et al.} KD&13.60\\
    &&            & \small~~+ Ours ($N=4$) \& Tung~\emph{et al.} KD&11.74\\
    \Xcline{1-5}{2pt}
    \end{tabular}}}
    \vspace{-0.2cm}
    \label{tab:quan2}
\end{table}

\begin{table}[t]
    \centering
      \caption{Experimental results on CelebA-HQ. }
    \vspace{-0.3cm}

   \renewcommand{\arraystretch}{1.1}
   \resizebox{\linewidth}{!}{\setlength{\tabcolsep}{1mm}{\begin{tabular}{cccllccccll}
    \Xcline{1-5}{2pt}
    \rule{0pt}{12pt}\multirow{1}{*}{\textbf{Model}}&\multirow{1}{*}{\textbf{Throughput}}& \multirow{1}{*}{\textbf{MACs (G)}} &\multirow{1}{*}{\textbf{KD Method}} & \multirow{1}{*}{\textbf{FID}} \\
    \Xcline{1-5}{2pt}
    \rule{0pt}{12pt}Teacher&1.47&248.68&Training without  KD &6.80\\ 
        \Xcline{1-5}{0.8pt}
    \rule{0pt}{12pt}\multirow{10}{*}{Student}&\multirow{10}{*}{3.52}&\multirow{10}{*}{121.80}&\rule{0pt}{12pt}Training without  KD&22.45\\
    \Xcline{4-5}{0.4pt}
    \Xcline{4-5}{0.4pt}
    &&                 & \rule{0pt}{10pt}\small~~+ Hinton~\emph{et al.} KD&13.55\\
    &&             & \small~~+ Zagoruyko~\emph{et al.} KD&15.36\\
    &&             & \small~~+ Tian~\emph{et al.} KD&15.35\\
    &&             & \small~~+ Tung~\emph{et al.} KD&13.36
    \\
    \Xcline{4-5}{0.4pt}
    &&          & \rule{0pt}{10pt}\small~~+ Ours ($N=4$) \& Hinton~\emph{et al.} KD& 8.92\\
    &&            & \small~~+ Ours ($N=4$) \& Zagoruyko~\emph{et al.} KD&  9.21\\
    &&            & \small~~+ Ours ($N=4$) \& Tian~\emph{et al.} KD& 9.56\\
    &&            & \small~~+ Ours ($N=4$) \& Tung~\emph{et al.} KD&7.62\\
    \Xcline{1-5}{2pt}
    \end{tabular}}}
        \vspace{-0.3cm}
    \label{tab:quan3}
\end{table}

\textbf{Qualitative Results } A qualitative comparison  between the students trained with and without O2MKD is shown in Figure~\ref{fig:lsun_qual}. On both unconditional generation and the text-to-image generation,
our approach surpasses the student baseline concerning the rationality, chromatic attributes, lucidity, and aesthetic qualities of the images.

\begin{figure}[t]
    \centering
    \begin{minipage}{0.48\linewidth}
        \includegraphics[width=\linewidth]{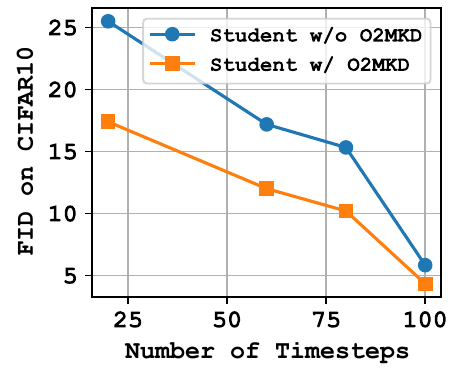}
        \vspace{-0.5cm}
        \caption{O2MKD with different DDIM sampling steps.}
        \vspace{-0.2cm}
        \label{fig:ddim1}
    \end{minipage}
    \hfill
    \begin{minipage}{0.48\linewidth}
    \vspace{-0.1cm}
        \includegraphics[width=\linewidth]{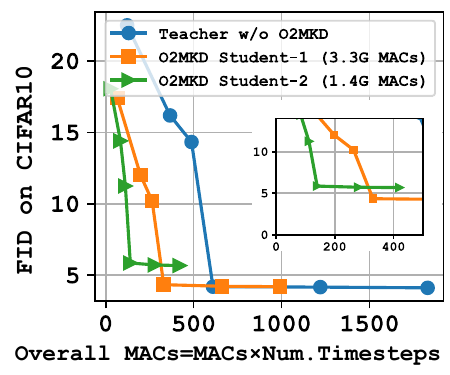}
        \vspace{-0.5cm}
        \caption{Comparison between O2MKD and DDIM.}
        \vspace{-0.2cm}
        \label{fig:ddim2}
    \end{minipage}
\end{figure}

\textbf{Compatibility with Fast Sampling Methods}
O2MKD aims to reduce the computational overhead of the UNet in a single denoising step, thus making it compatible with fast sampling methods that focus on reducing the number of sampling steps. As demonstrated in Figure~\ref{fig:ddim1}, for DDIM with varying numbers of sampling steps, O2MKD consistently yields a reduction in FID compared to students trained without knowledge distillation, suggesting that O2MKD is orthogonal to fast sampling methods.

Besides, we also compare the performance of DDIM and O2MKD in Figure~\ref{fig:ddim2}. It is observed that the student trained by O2MKD achieves lower FID than the teacher, especially when the overall computation is fewer than 500 MACs. We argue that this is because DDIM suffers from a significant performance drop when using an extremely small number of timesteps. In other words, O2MKD is a better choice to achieve acceleration when DDIM sampling steps have already been set as a small number. Besides, as shown in the subfigure of Figure~\ref{fig:ddim2}, O2MKD with a tiny student leads to higher FID than O2MKD with a relatively large teacher but smaller DDIM sampling steps, indicating that the cooperation between DDIM and O2MKD can achieve better performance than using only one of them.

\textbf{Comparison with Other Acceleration Methods} We also compare other diffusion acceleration methods including Diff-Pruning~\cite{structural_pruning_diffusion} and DeepCache~\cite{ma2024deepcache} in Table~\ref{tab:comparison}, which demonstrates that O2MKD achieves significant lower FID with the similar acceleration ratio, and O2MKD can be utilized with these methods to achieve better performance.

\begin{table}[t]
    \centering
        \caption{Experiments of text-to-image generation on MSCOCO 30K with Stable Diffusion v1.4 as the teacher and BK-SDM-Tiny~\cite{bksdm}, a pruned version of Stable Diffusion as the student.}
        \vspace{-0.32cm}
    \resizebox{\linewidth}{!}{\begin{tabular}{lcccc}
    \toprule
        Model & FID ($\downarrow$) & IS ($\uparrow$) & CLIP Score ($\uparrow$) & MACs ($\downarrow$)\\
       \midrule
       Stable Diffusion v1.4  &13.05&36.76&0.2958&9716  \\
       Student + Feature KD  & 17.12&30.09&0.2653&6373  \\
       Student + O2MKD ($N$=4) & 14.21&34.32&0.2871&6373 \\
       \bottomrule
    \end{tabular}}
\vspace{-0.1cm}
    \label{tab:text-to-image}
\end{table}

\begin{figure}[t]
    \centering
    \begin{minipage}[t]{0.50\linewidth} 
        \centering
        \captionof{table}{Comparison with other methods on CIFAR10.}
        \vspace{-0.3cm}
        \resizebox{\linewidth}{!}{
           \setlength{\tabcolsep}{0.3mm}{ \begin{tabular}{lcl}
            \toprule
                Method & MAC(G) & FID \\
            \midrule
                DeepCache & 3.5 & 5.73 \\
                DiffPruning & 3.4 & 5.29 \\
                O2MKD & 3.3 & 4.34 \\
                DeepCache+O2MKD & 2.8 & 4.51 \\
                DiffPruning+O2MKD & 3.4 & 4.17 \\
            \bottomrule
            \vspace{-0.5cm}
            \end{tabular}}
        }
        \label{tab:comparison}
    \end{minipage}%
    \hfill
    \begin{minipage}[t]{0.49\linewidth} 
        \centering
        \vspace{0pt} 
        \captionof{table}{O2MKD with model merge on CIFAR10.}
        \vspace{-0.2cm}
        \resizebox{\linewidth}{!}{
            \setlength{\tabcolsep}{0.25mm}{ \begin{tabular}{l|cc}
            \toprule
                Model & $N$=4 & $N$=8 \\
            \midrule
                O2MKD before merge & 4.62 & 4.34 \\
                O2MKD after merge & 4.75 & 4.42 \\
            \midrule
                student w/ trad. KD & 5.32 & 5.32 \\
                student w/o KD & 5.84 & 5.84 \\
            \bottomrule
 
            \end{tabular}}
        }           \vspace{-0.5cm}
        \label{tab:merge}
    \end{minipage}
\end{figure}

\section{Discussion}

\subsection{Memory Footprint Analysis}  The peak memory usage of a neural network can be roughly estimated as the sum of the memory footprint of its parameters and features. In O2MKD, this can be expressed as $P \times N + F \times B$, where $P, N, F, B$ represent the number of parameters, the number of students, the memory footprint of the largest feature map, and the batch size, respectively. As shown in Figure~\ref{fig:memory}, during generating images with 256x256 resolutions, the memory footprint of O2MKD is significantly smaller than the teacher, and its additional memory footprint compared with traditional KD is smaller enough to be ignored. Note that the additional memory from using multiple students in O2MKD is more ignorable when generating higher-resolution images since in these settings the feature map of images dominates the overall memory usage.

\textbf{Reducing Memory Footprint with Model Merge} Model merging is a technique that can merge the parameters of multiple models into one model. We find that it is possible to merge multiple students in O2MKD into one student to eliminate the additional memory footprint overhead~\cite{DBLP:conf/iclr/IlharcoRWSHF23}. As shown in Table~\ref{tab:merge}, model merge on O2MKD leads to 0.13 and 0.08 FID improvements, which are still significantly lower than the student trained without KD and the student with traditional KD, indicating that the additional memory problem from using multiple students in O2MKD may be solved by model merge.
\subsection{Ablation Study}

\subsubsection{Influence of the Number of Students ($N$)}

\begin{figure}[t]
    \centering
    \includegraphics[width=\linewidth]{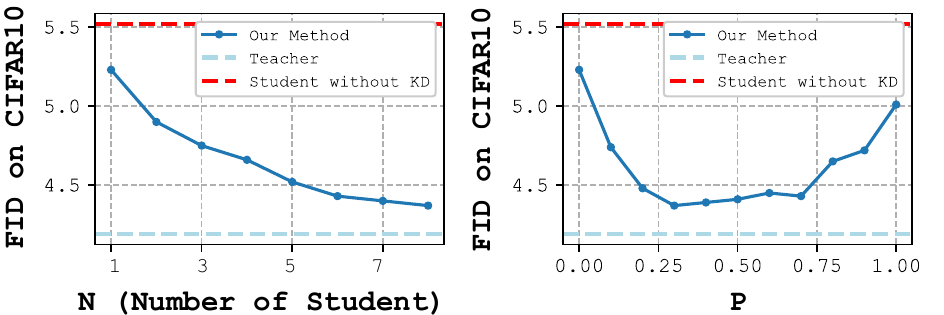}
    \caption{Influence from the number of students $N$ and the possibility of training on all timesteps $p$ on CIFAR10.}
    \label{fig:hyper}
\end{figure}

O2MKD is designed to distill knowledge from a teacher model to a group of $N$ student models. A larger $N$ implies that each student is trained to mimic the teacher's knowledge for a narrower range of timesteps, potentially leading to improved image synthesis performance. While the increase in $N$ doesn't result in additional computational overhead during inference, it does introduce more parameters and increased memory usage during inference. In this subsection, we provide a detailed analysis of the influence of $N$. As depicted in Figure~\ref{fig:hyper}(a), a larger $N$ generally correlates with a lower FID, signifying enhanced performance. For instance, using two students yields a 0.33 FID reduction compared to using a single student. However, when the number of students becomes considerably larger, such as eight students versus seven students, the additional benefits become marginal, with only a 0.03 FID reduction. This suggests that the advantages of employing more students diminish as $N$ increases significantly. Please note that some data points such as $N$=1, $P$=0, and $P$=1 have bad performance because in these settings O2MKD degenerates into the traditional KD.


\begin{figure}[t]
    \centering
    \begin{minipage}{\linewidth}
        \centering
        \begin{minipage}[t]{0.57\linewidth}
            \centering
            \vspace{-3.cm}
            \resizebox{\linewidth}{!}{
                \begin{tabular}{l|rrr}
                     \toprule
                Hyper-parameter $p$ & $p=1$ & $p=0.6$ \\
                \midrule
                Diffusion Loss & 283 & 132 \\
                KD Loss & 0.40 & 0.31 \\
                FID$^{1}$ & 432.4 & 5.27 \\
                FID$^{2}$ & 4.52 & 4.33 \\
                \bottomrule
            \end{tabular}
            }
            \vspace{-0.3cm}
            \captionof{table}{Students trained for $[0, 250)$ timesteps in O2MKD. FID$^1$: Using the student for generation at all time steps.
            FID$^2$: Using the student for generation at [0, 250) time steps and using the teacher for other time steps.    }
            \vspace{-0.3cm}
            \label{tab:p}
        \end{minipage}
        \hfill
        \begin{minipage}[t]{0.37\linewidth}
            \centering
            \includegraphics[width=1.05\linewidth]{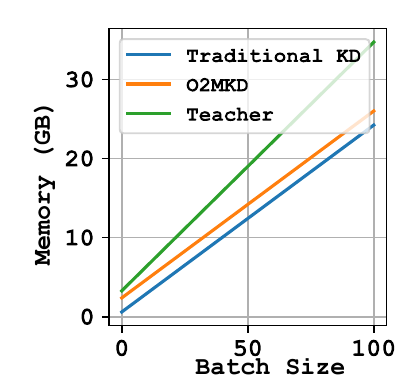}
            \captionof{figure}{Memory footprint comparison on LSUN Church.}
            \vspace{-0.3cm}
            \label{fig:memory}
        \end{minipage}
    \end{minipage}
\end{figure}

\subsection{Influence from the Probability $p$}

A straightforward implementation of O2MKD is to directly train each student with timesteps belonging to
their designated time steps (\emph{i.e.,} $p=1$).
However, as shown in Figure~\ref{fig:hyper}(b), the aforementioned intuitive implementation ($p=1$)  leads to 
significant performance drop compared with O2MKD with $p=0.6$.
To analyze this phenomenon, we study the performance of a single student in the student group with different $p$ in Table~\ref{tab:p}.
These two students are specifically trained for timesteps $t\in [0, 250)$.
It is observed that:
(1) The student trained with $p=1$ exhibits considerably larger KD loss and diffusion loss compared to the student trained with $p=0.6$, indicating bad convergence.
(2) When employing the student for all time steps, the student trained with $p=0.6$ performs effectively while the student trained with $p=1$ fails.
(3) When employing the student for $[250, 1000)$ and the teacher for the other time steps, the student trained by $p=1$ still
leads to a higher FID, but its disadvantage to the student trained by $p=0.6$ is not that significant.
These observations signify that training the student with only its corresponding time steps makes the student lose the knowledge of the other time steps. In contrast, training the student beyond its corresponding time steps with a suitable probability can notably enhance their performance.
We posit that this training approach enables the student to learn from more data, which makes a good trade-off between general knowledge and specific knowledge.




\subsection{O2MKD Beyond Model Compression}
We further investigate the possibility of employing O2MKD as a general training technique for the targets beyond compression. Concretely, in this setting, we set the architecture of the student models to be identical to that of the teacher model, initialize them with the parameters of the teacher model, and train them with O2MKD. Our experimental results demonstrate that in this way we reduce FID from \textbf{4.19} to \textbf{3.78} on CIFAR10, and \textbf{10.60} to \textbf{8.75} on LSUN Church, indicating that O2MKD can also be utilized to improve the generation quality in the non-compression setting.

\subsection{Knowledge Distillation Provides Stable Supervision}
\begin{figure}[t]
\includegraphics[width=\linewidth]{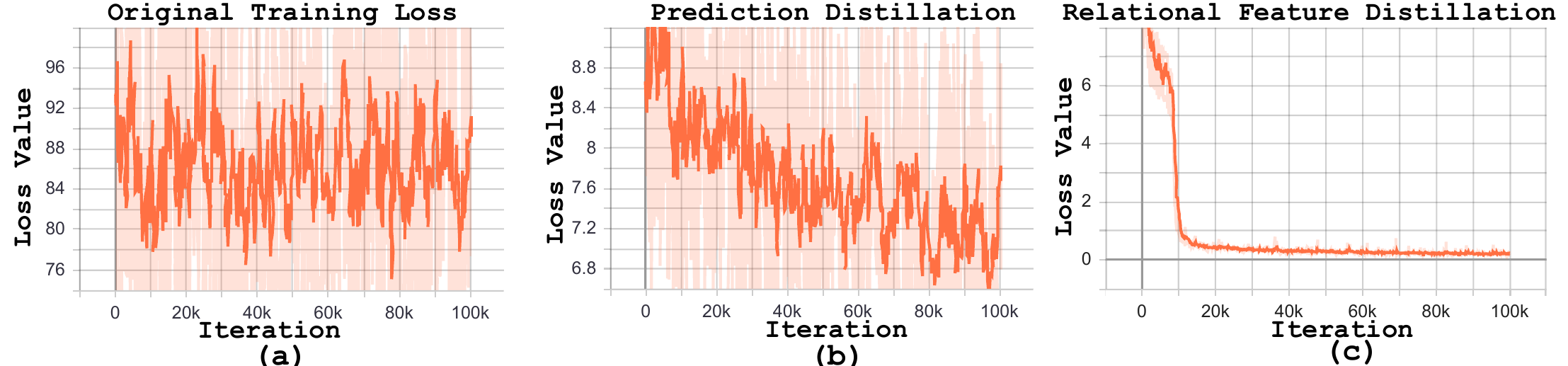}

\vspace{-0.2cm}
    \caption{The training loss of students on CIFAR10 with exponential moving average:
    (a) the original training loss of diffusion models, (b) the training loss of prediction distillation~\cite{distill_hinton}, (c) the training loss of relational knowledge distillation~\cite{relational_kd2}.
    }
    \vspace{-0.4cm}
    \label{fig:training_loss}
\end{figure}

The visualization of the training loss, including the original training loss, prediction-based KD loss, and relational feature distillation loss, is depicted in Figure~\ref*{fig:training_loss}. Several key observations can be made:
(1) The original training loss plays a pivotal role in training student diffusion models, but its value exhibits significant instability and does not demonstrate a clear reduction.
(2) Similarly, the prediction-based KD loss also exhibits instability, but it has undergone a notable reduction during the training process.
(3) In contrast, the relational feature distillation loss displays a remarkably stable trend and experiences significant reduction throughout the training phase.
These observations suggest the following:
(1) Both KD methods have clear reductions, indicating that KD provides valuable supervision that facilitates model training, which improves the performance of student models.
(2) Relational feature distillation offers more stable supervision compared to prediction-based KD and the original training loss. This is further supported by the fact that relational feature distillation methods result in the lowest FID in all experimental settings.
We posit that the instability observed in prediction-based KD and the original training loss may be attributed to the significant outliers in the predicted noise. Instead, relational feature distillation captures the similarities between feature values at different positions, not just their individual values.

\subsection{O2MKD with Non-Uniform Timestep Ranges}
In O2MKD, all the students are uniformly assigned to the same number of timesteps ($\frac{T}{N}$). Here we study O2MKD with non-uniform timestep ranges with the following three schemes on CIFAR10: (1) Scheme-A: More students are assigned to larger (noisier) timesteps, which achieves 4.52 FID. (B) Scheme-B: More students are assigned to smaller timesteps, which achieves 4.23 FID. (C) Uniform Scheme: Students are assigned to the same number of timesteps, which achieves 4.33 FID. These results demonstrate that dividing the timesteps in a non-uniform manner, \emph{i.e.} applying more students to learn teacher knowledge in smaller timesteps, may lead to better performance. 

\section{Conclusion}
This paper introduces one-to-many knowledge distillation (O2MKD), a novel KD framework to accelerate diffusion models. O2MKD leverages the observation that diffusion models exhibit different behaviors at different timesteps, and proposes to distill knowledge from a teacher across all time steps to multiple students, each dedicated to continuous and non-overlapping time steps. This can be thought of as transferring knowledge from a general model to several domain experts. Our experiments demonstrate the effectiveness of O2MKD across various training settings, and it can be seamlessly integrated with existing knowledge distillation methods and fast sampling techniques. Furthermore, we discover that the supervision from KD exhibits significantly greater stability compared to supervision from traditional training loss of diffusion models, which may offer a novel viewpoint for the development of diffusion training methodologies.

\bibliography{aaai25}

\begin{thebibliography}{36}
\providecommand{\natexlab}[1]{#1}

\bibitem[{Balaji et~al.(2022)Balaji, Nah, Huang, Vahdat, Song, Kreis, Aittala, Aila, Laine, Catanzaro et~al.}]{balaji2022ediffi}
Balaji, Y.; Nah, S.; Huang, X.; Vahdat, A.; Song, J.; Kreis, K.; Aittala, M.; Aila, T.; Laine, S.; Catanzaro, B.; et~al. 2022.
\newblock ediffi: Text-to-image diffusion models with an ensemble of expert denoisers.
\newblock \emph{arXiv preprint arXiv:2211.01324}.

\bibitem[{Bo-Kyeong et~al.(2023)Bo-Kyeong, Hyoung-Kyu, Thibault, and Shinkook}]{bksdm}
Bo-Kyeong, K.; Hyoung-Kyu, S.; Thibault, C.; and Shinkook, C. 2023.
\newblock Bk-sdm: A lightweight, fast, and cheap version of stable diffusion.
\newblock \emph{arXiv preprint arXiv:2305.15798v3}.

\bibitem[{Fang, Ma, and Wang(2023)}]{structural_pruning_diffusion}
Fang, G.; Ma, X.; and Wang, X. 2023.
\newblock Structural Pruning for Diffusion Models.
\newblock \emph{arXiv preprint arXiv:2305.10924}.

\bibitem[{Gao et~al.(2023)Gao, Liu, Zeng, Xu, Li, Luo, Liu, Zhen, and Zhang}]{diffusion_sr}
Gao, S.; Liu, X.; Zeng, B.; Xu, S.; Li, Y.; Luo, X.; Liu, J.; Zhen, X.; and Zhang, B. 2023.
\newblock Implicit diffusion models for continuous super-resolution.
\newblock In \emph{Proceedings of the IEEE/CVF Conference on Computer Vision and Pattern Recognition}, 10021--10030.

\bibitem[{Goodfellow et~al.(2014)Goodfellow, Pouget-Abadie, Mirza, Xu, Warde-Farley, Ozair, Courville, and Bengio}]{gan}
Goodfellow, I.; Pouget-Abadie, J.; Mirza, M.; Xu, B.; Warde-Farley, D.; Ozair, S.; Courville, A.; and Bengio, Y. 2014.
\newblock Generative adversarial nets.
\newblock \emph{Advances in neural information processing systems}, 27.

\bibitem[{Hinton, Vinyals, and Dean(2014)}]{distill_hinton}
Hinton, G.; Vinyals, O.; and Dean, J. 2014.
\newblock Distilling the knowledge in a neural network.
\newblock In \emph{NeurIPS}.

\bibitem[{Ho et~al.(2022)Ho, Chan, Saharia, Whang, Gao, Gritsenko, Kingma, Poole, Norouzi, Fleet et~al.}]{diffusion_video_1}
Ho, J.; Chan, W.; Saharia, C.; Whang, J.; Gao, R.; Gritsenko, A.; Kingma, D.~P.; Poole, B.; Norouzi, M.; Fleet, D.~J.; et~al. 2022.
\newblock Imagen video: High definition video generation with diffusion models.
\newblock \emph{arXiv preprint arXiv:2210.02303}.

\bibitem[{Ho, Jain, and Abbeel(2020)}]{ddpm}
Ho, J.; Jain, A.; and Abbeel, P. 2020.
\newblock Denoising diffusion probabilistic models.
\newblock \emph{Advances in neural information processing systems}, 33: 6840--6851.

\bibitem[{Ilharco et~al.(2022)Ilharco, Ribeiro, Wortsman, Gururangan, Schmidt, Hajishirzi, and Farhadi}]{ilharco2022editing}
Ilharco, G.; Ribeiro, M.~T.; Wortsman, M.; Gururangan, S.; Schmidt, L.; Hajishirzi, H.; and Farhadi, A. 2022.
\newblock Editing models with task arithmetic.
\newblock \emph{arXiv preprint arXiv:2212.04089}.

\bibitem[{Ilharco et~al.(2023)Ilharco, Ribeiro, Wortsman, Schmidt, Hajishirzi, and Farhadi}]{DBLP:conf/iclr/IlharcoRWSHF23}
Ilharco, G.; Ribeiro, M.~T.; Wortsman, M.; Schmidt, L.; Hajishirzi, H.; and Farhadi, A. 2023.
\newblock Editing models with task arithmetic.
\newblock In \emph{The Eleventh International Conference on Learning Representations, {ICLR} 2023, Kigali, Rwanda, May 1-5, 2023}. OpenReview.net.

\bibitem[{Isola et~al.(2017)Isola, Zhu, Zhou, and Efros}]{pix2pix}
Isola, P.; Zhu, J.-Y.; Zhou, T.; and Efros, A.~A. 2017.
\newblock Image-to-image translation with conditional adversarial networks.
\newblock In \emph{Proceedings of the IEEE conference on computer vision and pattern recognition}, 1125--1134.

\bibitem[{Karras et~al.(2017)Karras, Aila, Laine, and Lehtinen}]{karras2017progressive}
Karras, T.; Aila, T.; Laine, S.; and Lehtinen, J. 2017.
\newblock Progressive growing of gans for improved quality, stability, and variation.
\newblock \emph{arXiv preprint arXiv:1710.10196}.

\bibitem[{Kingma and Welling(2013)}]{vae}
Kingma, D.~P.; and Welling, M. 2013.
\newblock Auto-encoding variational bayes.
\newblock \emph{arXiv preprint arXiv:1312.6114}.

\bibitem[{Krizhevsky and Hinton(2009)}]{cifar}
Krizhevsky, A.; and Hinton, G. 2009.
\newblock Learning multiple layers of features from tiny images.
\newblock Technical report, Citeseer.

\bibitem[{Lee et~al.(2023)Lee, Kim, Go, Jeong, Oh, and Choi}]{Lee2023MultiArchitectureMD}
Lee, Y.; Kim, J.-Y.; Go, H.; Jeong, M.; Oh, S.; and Choi, S. 2023.
\newblock Multi-Architecture Multi-Expert Diffusion Models.
\newblock \emph{ArXiv}, abs/2306.04990.

\bibitem[{Li et~al.(2023)Li, Lian, Liu, Yang, Dong, Kang, Zhang, and Keutzer}]{diffusion_quan1}
Li, X.; Lian, L.; Liu, Y.; Yang, H.; Dong, Z.; Kang, D.; Zhang, S.; and Keutzer, K. 2023.
\newblock Q-diffusion: Quantizing diffusion models.
\newblock \emph{arXiv preprint arXiv:2302.04304}.

\bibitem[{Lin et~al.(2014)Lin, Maire, Belongie, Hays, Perona, Ramanan, Doll{\'a}r, and Zitnick}]{mscoco}
Lin, T.-Y.; Maire, M.; Belongie, S.; Hays, J.; Perona, P.; Ramanan, D.; Doll{\'a}r, P.; and Zitnick, C.~L. 2014.
\newblock Microsoft coco: Common objects in context.
\newblock In \emph{European conference on computer vision}, 740--755. Springer.

\bibitem[{Liu et~al.(2023)Liu, Ning, Lin, Yang, and Wang}]{diffusion_mos_dpm}
Liu, E.; Ning, X.; Lin, Z.; Yang, H.; and Wang, Y. 2023.
\newblock OMS-DPM: Optimizing the Model Schedule for Diffusion Probabilistic Models.
\newblock \emph{arXiv preprint arXiv:2306.08860}.

\bibitem[{Luhman and Luhman(2021)}]{diffusion_distillation_3}
Luhman, E.; and Luhman, T. 2021.
\newblock Knowledge distillation in iterative generative models for improved sampling speed.
\newblock \emph{arXiv preprint arXiv:2101.02388}.

\bibitem[{Ma, Fang, and Wang(2024)}]{ma2024deepcache}
Ma, X.; Fang, G.; and Wang, X. 2024.
\newblock Deepcache: Accelerating diffusion models for free.
\newblock In \emph{Proceedings of the IEEE/CVF Conference on Computer Vision and Pattern Recognition}, 15762--15772.

\bibitem[{Meng et~al.(2023)Meng, Rombach, Gao, Kingma, Ermon, Ho, and Salimans}]{diffusion_distillation_1}
Meng, C.; Rombach, R.; Gao, R.; Kingma, D.; Ermon, S.; Ho, J.; and Salimans, T. 2023.
\newblock On distillation of guided diffusion models.
\newblock In \emph{Proceedings of the IEEE/CVF Conference on Computer Vision and Pattern Recognition}, 14297--14306.

\bibitem[{Nichol and Dhariwal(2021)}]{iddpm}
Nichol, A.~Q.; and Dhariwal, P. 2021.
\newblock Improved denoising diffusion probabilistic models.
\newblock In \emph{International Conference on Machine Learning}, 8162--8171. PMLR.

\bibitem[{Poole et~al.(2022)Poole, Jain, Barron, and Mildenhall}]{dreamfusion}
Poole, B.; Jain, A.; Barron, J.~T.; and Mildenhall, B. 2022.
\newblock Dreamfusion: Text-to-3d using 2d diffusion.
\newblock \emph{arXiv preprint arXiv:2209.14988}.

\bibitem[{Rombach et~al.(2022)Rombach, Blattmann, Lorenz, Esser, and Ommer}]{latent_diffusion}
Rombach, R.; Blattmann, A.; Lorenz, D.; Esser, P.; and Ommer, B. 2022.
\newblock High-resolution image synthesis with latent diffusion models.
\newblock In \emph{Proceedings of the IEEE/CVF conference on computer vision and pattern recognition}, 10684--10695.

\bibitem[{Romero et~al.(2015)Romero, Ballas, Kahou, Chassang, Gatta, and Bengio}]{fitnets}
Romero, A.; Ballas, N.; Kahou, S.~E.; Chassang, A.; Gatta, C.; and Bengio, Y. 2015.
\newblock Fitnets: Hints for thin deep nets.
\newblock In \emph{ICLR}.

\bibitem[{Salimans and Ho(2022)}]{diffusion_distillation_2}
Salimans, T.; and Ho, J. 2022.
\newblock Progressive distillation for fast sampling of diffusion models.
\newblock \emph{arXiv preprint arXiv:2202.00512}.

\bibitem[{Sasaki, Willcocks, and Breckon(2021)}]{diffusion_i2im1}
Sasaki, H.; Willcocks, C.~G.; and Breckon, T.~P. 2021.
\newblock Unit-ddpm: Unpaired image translation with denoising diffusion probabilistic models.
\newblock \emph{arXiv preprint arXiv:2104.05358}.

\bibitem[{Shang et~al.(2023)Shang, Yuan, Xie, Wu, and Yan}]{diffusion_quan2}
Shang, Y.; Yuan, Z.; Xie, B.; Wu, B.; and Yan, Y. 2023.
\newblock Post-training quantization on diffusion models.
\newblock In \emph{Proceedings of the IEEE/CVF Conference on Computer Vision and Pattern Recognition}, 1972--1981.

\bibitem[{Song, Meng, and Ermon(2020)}]{ddim}
Song, J.; Meng, C.; and Ermon, S. 2020.
\newblock Denoising diffusion implicit models.
\newblock \emph{arXiv preprint arXiv:2010.02502}.

\bibitem[{Sun et~al.(2022)Sun, Chen, Wang, Ye, Feng, and Chen}]{diffusion_distillation_4}
Sun, W.; Chen, D.; Wang, C.; Ye, D.; Feng, Y.; and Chen, C. 2022.
\newblock Accelerating diffusion sampling with classifier-based feature distillation.
\newblock \emph{arXiv preprint arXiv:2211.12039}.

\bibitem[{Tung and Mori(2019)}]{relational_kd2}
Tung, F.; and Mori, G. 2019.
\newblock Similarity-preserving knowledge distillation.
\newblock In \emph{Proceedings of the IEEE International Conference on Computer Vision}, 1365--1374.

\bibitem[{Yang, Srivastava, and Mandt(2022)}]{diffusion_video_2}
Yang, R.; Srivastava, P.; and Mandt, S. 2022.
\newblock Diffusion probabilistic modeling for video generation.
\newblock \emph{arXiv preprint arXiv:2203.09481}.

\bibitem[{Yu et~al.(2015)Yu, Seff, Zhang, Song, Funkhouser, and Xiao}]{lsun}
Yu, F.; Seff, A.; Zhang, Y.; Song, S.; Funkhouser, T.; and Xiao, J. 2015.
\newblock Lsun: Construction of a large-scale image dataset using deep learning with humans in the loop.
\newblock \emph{arXiv preprint arXiv:1506.03365}.

\bibitem[{Zagoruyko and Komodakis(2017)}]{attentiondistillation}
Zagoruyko, S.; and Komodakis, N. 2017.
\newblock Paying more attention to attention: Improving the performance of convolutional neural networks via attention transfer.
\newblock In \emph{ICLR}.

\bibitem[{Zhu et~al.(2017)Zhu, Park, Isola, and Efros}]{cyclegan}
Zhu, J.-Y.; Park, T.; Isola, P.; and Efros, A.~A. 2017.
\newblock Unpaired image-to-image translation using cycle-consistent adversarial networks.
\newblock In \emph{Proceedings of the IEEE international conference on computer vision}, 2223--2232.

\bibitem[{Zhu et~al.(2023)Zhu, Zhang, Liang, Cao, Wen, Timofte, and Van~Gool}]{diffusion_restoration}
Zhu, Y.; Zhang, K.; Liang, J.; Cao, J.; Wen, B.; Timofte, R.; and Van~Gool, L. 2023.
\newblock Denoising Diffusion Models for Plug-and-Play Image Restoration.
\newblock In \emph{Proceedings of the IEEE/CVF Conference on Computer Vision and Pattern Recognition}, 1219--1229.

\end{thebibliography}

\end{document}